\documentclass[letterpaper, 10 pt, conference]{ieeeconf}  

\usepackage{colortbl}
\usepackage{graphicx,xcolor}

\usepackage{bm}
\usepackage{subfig}
\usepackage{amssymb}
\usepackage{mathtools}
\usepackage{array,multirow}
\usepackage{balance}

\definecolor{bed}{RGB}{17, 128, 128}
\definecolor{books}{RGB}{247, 53, 58}
\definecolor{ceiling}{RGB}{101, 27, 201}
\definecolor{chair}{RGB}{52, 62, 246}
\definecolor{floor}{RGB}{230, 230, 237}
\definecolor{furniture}{RGB}{252, 71, 38}
\definecolor{objects}{RGB}{252, 32, 128}
\definecolor{painting}{RGB}{51, 54, 148}
\definecolor{sofa}{RGB}{221, 180, 143}
\definecolor{table}{RGB}{65, 248, 68}
\definecolor{tv}{RGB}{254, 213, 48}
\definecolor{wall}{RGB}{150, 150, 150}
\definecolor{window}{RGB}{45, 255, 254}

\definecolor{slam}{RGB}{201, 213, 255}
\definecolor{seg}{RGB}{255, 231, 190}
\definecolor{rec}{RGB}{224, 255, 215}

\newcommand{\argmax}{\mathop{\rm arg~max}\limits}

\newcommand{\bhline}[1]{\noalign{\hrule height #1}}

\IEEEoverridecommandlockouts

\title{\LARGE \bf
Fast and Accurate Semantic Mapping through Geometric-based Incremental Segmentation
}

\author{Yoshikatsu Nakajima$^{1}$, Keisuke Tateno$^{2}$, Federico Tombari$^{2}$ and Hideo Saito$^{1}$
\thanks{$^{1}$Department of Science and Technology, Keio University, Kanagawa, (Japan)
        {\tt\small \{nakajima, saito\}@hvrl.ics.keio.ac.jp}}%
\thanks{$^{2}$Chair for Computer Aided Medical Procedures (CAMP), TU Munich, Boltzmannstr. 3, 85748 Munich (Germany)
        {\tt\small \{tateno, tombari\}@in.tum.de}}%
}

\begin{document}

\maketitle
\thispagestyle{empty}
\pagestyle{empty}

\begin{abstract}

We propose an efficient and scalable method for incrementally building a dense, semantically annotated 3D map in real-time.
The proposed method assigns class probabilities to each region, not each element (e.g., surfel and voxel), of the 3D map which is built up through a robust SLAM framework and incrementally segmented with a geometric-based segmentation method.
Differently from all other approaches, our method has a capability of running at over 30Hz while performing all processing components, including SLAM, segmentation, 2D recognition, and updating class probabilities of each segmentation label at every incoming frame, thanks to the high efficiency that characterizes the computationally intensive stages of our framework.
By utilizing a specifically designed CNN to improve the frame-wise segmentation result, we can also achieve high accuracy.
We validate our method on the NYUv2 dataset by comparing with the state of the art in terms of accuracy and computational efficiency, and by means of an analysis in terms of time and space complexity.

\end{abstract}

\section{INTRODUCTION}

The task of incrementally building a semantically annotated 3D map is a challenging research topic for both the robotics and computer vision communities. It has a wide range of applications including autonomous grasping and manipulation of objects, scene understanding, robotics navigation and augmented reality. For this reason, a valuable research effort is currently undergoing in literature with the aim of developing efficient systems that can scale up to mobile/embedded architectures while being robust enough to generalize to unseen environments.

\begin{figure}[t]
\begin{center}
    \includegraphics[width=1.0\hsize]{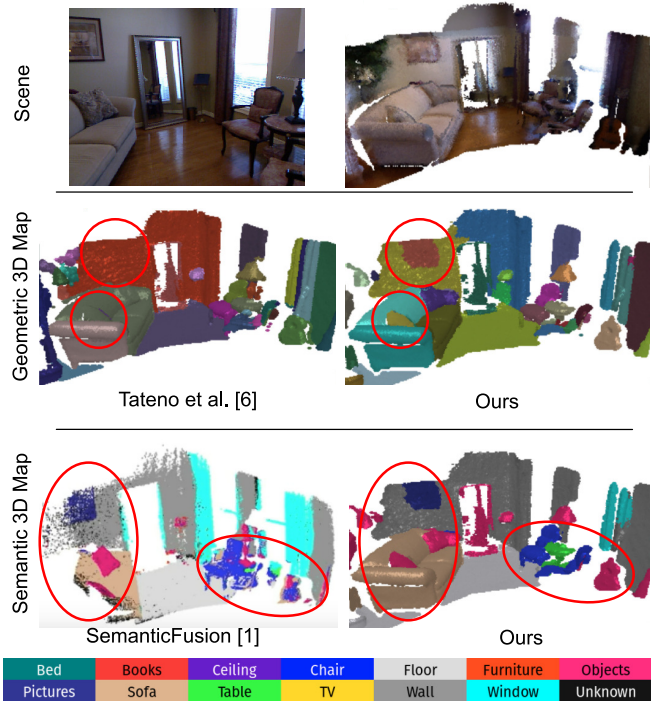}
    \caption{Our method achieves accurate semantic mapping (comparable to the state of the art \cite{mccormac2017semanticfusion}, bottom row) while being more efficient and scalable. It relies on the geometric segmentation that takes into account semantic information and thus providing meaningful segments than the method of Tateno et al. \cite{tateno2015real} (middle row).
 \label{fig:teaser}}
\end{center}
\end{figure}

Motivated by the recent developments of deep learning and Convolutional Neural Networks (CNNs) for 3D data, recent methods have mostly focused on increasing the accuracy of the semantic segmentation map \cite{mccormac2017semanticfusion,li2016semi,yang2017semantic}. At the same time, they still face the critical issue of yielding real-time performance, since such systems are built on a set of computationally demanding processing stages, including 3D reconstruction, camera pose estimation and CNN-based semantic segmentation. This becomes even more relevant with regards to embedded and mobile architectures that are typically employed for the aforementioned applications of robotics navigation/grasping and augmented reality. 

To achieve real-time performance, some of these methods suggested to only extract semantic information on a subset of the input frames. 
For example, the methods proposed by Hermans et al. \cite{hermans2014dense} and McCormac et al. (SemanticFusion) \cite{mccormac2017semanticfusion} achieved, respectively, an output frame-rate of 4Hz and 25.3Hz, by running semantic segmentation, respectively, every 6 and every 10 frames.
While such frame skipping strategy can improve run-time performance, it limits their range of application, since it tends to bring in inaccuracies under fast camera motions.

In this paper, we propose a novel incremental semantic mapping approach that aims at overcoming such issues by yielding highly accurate semantic scene reconstruction (see bottom row of Fig. \ref{fig:teaser}) in real-time. The framework relies on effectively combining a reliable camera pose tracking (InfiniTAM v3 \cite{InfiniTAM_V3_Report_2017}), an incremental segmentation approach \cite{tateno2015real}, and an efficient CNN-based semantic segmentation method. In particular, the 3D map of the scene is built through the fast and robust surfel-based SLAM approach in \cite{keller2013real}, and geometric segmentation labels are assigned to each surfel based on the approach of \cite{tateno2015real}. Class probabilities of each label are updated through a specifically designed CNN. 

We introduce a new probabilistic strategy to deal with one of the most delicate stages, i.e. class probability assignment. According to this strategy, and in contrast to conventional semantic mapping methods which assign class probabilities to each surfel \cite{hermans2014dense,mccormac2017semanticfusion,li2016semi}, we assign class probabilities to each segment. This reduces notably the time complexity since at each new frame probability distributions need to be updated for those segments which are visible on the image plane from the current camera pose, in contrast to conventional methods which need to update such probabilities for all surfels on the image plane.
This strategy also reduces notably the space complexity since probability distributions need to be stored only at each segment rather than each surfel.

In return, the semantic information also improves the geometric-based segmentation from \cite{tateno2015real}. By taking into account semantic information, it provides additional edges that better represent the semantic structure of the scene, hence allowing to obtain accurate segment regions (see middle row of Fig. \ref{fig:teaser}). Since smoothing of semantic labels is carried out at the geometric fusion stage, this allows us to utilize a CNN with a low resolution (i.e. $40\times30$) output, with a forward pass requiring only 19ms on an off-the-shelf GPU (i.e., a GeForce GTX 1080).

The overall framework is capable of working in real-time on off-the-shelf architectures, while the requiring a low computational complexity with respect to state of the art. 
In addition, differently from other methods such as \cite{hermans2014dense,mccormac2017semanticfusion,li2016semi,yang2017semantic,kundu2014joint}, our approach does not require any post-processing based on, e.g., Conditional Random Field, to refine the output of the semantic mapping.  
We demonstrate the effectiveness and efficiency of our approach on a common benchmark, i.e. the NYUv2 dataset \cite{silberman2012indoor}, reporting comparable accuracy than the state-of-the-art approaches while being notably faster and scaling better in terms of memory requirements. In addition, we also report an analysis in terms of time and space complexity of our method, demonstrating its advantages with respect to conventional approaches.

\begin{figure*}[t]
\begin{center}
    \includegraphics[width=1.0\hsize]{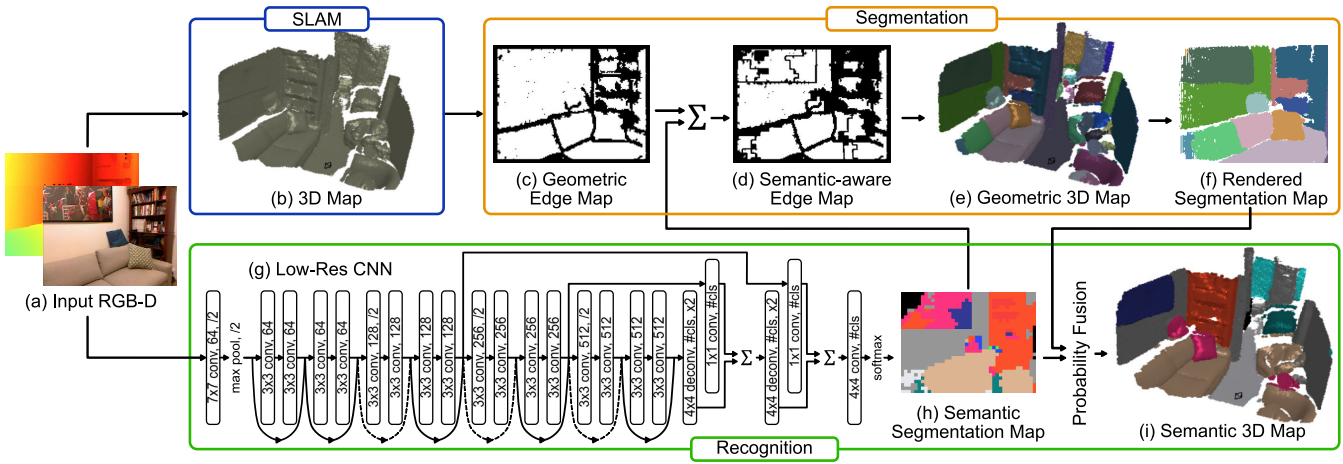}
    \caption{Flow of the proposed framework. Efficient CNN-based semantic segmentation is exploited to refine the geometric edges on frame-wise segmentation, then it is efficiently fused in the SLAM-based model using the rendered viewpoint according to the estimated camera pose. \label{fig:flow}}
\end{center}
\end{figure*}

\section{RELATED WORK}

\subsection{Semantic mapping}
Related work aimed at incrementally computing a semantic 3D map of the environment are mostly build on top of the following three main stages: (i) frame-wise segmentation to estimate the per-pixel class probability of the input frame, (ii) 2D-3D label transfer to fuse the 2D semantic segmentation labels to the 3D map; and, (iii) 3D refinement to denoise the class probabilities of the 3D map \cite{hermans2014dense,mccormac2017semanticfusion,li2016semi,yang2017semantic,vineet2015incremental,kundu2014joint}.
Notably, \cite{hermans2014dense} employed Random Decision Forests (RDF), a Bayesian framework and Conditional Random Field (CRF) respectively to carry out the three above-mentioned stages.

Since the CRF works on each element of the 3D map reconstructed via SLAM, it is effective in refining the semantic model and obtain high accuracy. Nevertheless, it is computationally heavy, as it requires 400 to 1800ms just for the CRF stage, yielding a frame-rate of 3.9 to 4.6Hz even if the method computes the RDF once every 6 input frames and the CRF once every 30 frames.
SemanticFusion \cite{mccormac2017semanticfusion} employs the CNN model proposed by Noh et al. \cite{noh2015learning} for 2D semantic segmentation, a Bayesian framework for 2D-3D label transfer, and a CRF for 3D refinement.
By using a CNN to carry out semantic segmentation of each input frame, the method can achieve a better runtime performance. However, the CNN still requires 51.2ms and the Bayesian update scheme requires a further 41.1ms, eventually running at 25.3Hz by applying these stages once every 10 input frames.

Other related works include \cite{sengupta2013urban,koppula2011semantic,zhao2016building} that aim at building a semantic 3D map, although not incrementally.
\cite{sengupta2013urban} firstly builds a 3D map of a scene through RGB-D SLAM framework, then assigns class probabilities to each point of the 3D map by means of a Dense CRF.
\cite{koppula2011semantic} exploits relational information derived from the full-scene 3D map for object labeling relying on a Markov-Random-Field (MRF)-based model.

In addition, several methods for recognizing only a part of the 3D map without making a dense semantic 3D map have been proposed \cite{salas2013slam++,bowman2017probabilistic,galvez2016real,fioraio2013joint}.
SLAM++ \cite{salas2013slam++} maps indoor scenes at the level of semantically defined objects.
Bowman et al. \cite{bowman2017probabilistic} improved the RGB SLAM performance in terms of camera pose and scale estimation by utilizing not only low-level geometric features such as points, lines, and planes but also detected objects as landmarks.

\subsection{2D semantic segmentation}
Several CNN models \cite{long2015fully,badrinarayanan2017segnet,noh2015learning,chen2017rethinking} for semantic segmentation have been proposed, sometimes yielding impressive results.
To achieve a highly precise semantic segmentation map, such methods aim at exploiting global information and context to improve the features extracted by the convolutional layers. In particular, Fully Convolutional Network (FCN) \cite{long2015fully} proposed a skip architecture that combines semantic information from a deep layer with appearance information from a shallow layer to perform accurate and detailed segmentation.

\subsection{3D geometric segmentation}
On the other hand, 3D geometric segmentation algorithms have been developed, to extract geometrically separated segments from 3D data by unsupervised fashion. Real-time segmentation for depth map has been investigated by the works of Uckermann et. al. \cite{Uckermann2013, Uckermann2012}, Pieropan et al. \cite{Pieropan2014} and Abramov et al. \cite{Abramov2012}. As a consequence, in addition to frame-wise segmentation, \cite{Finman2013, tateno2015real} has addressed the problem of real-time geometric segmentation for 3D point cloud or 3D mesh reconstructed via dense SLAM by incremental approach.

\section{METHOD}
Fig. \ref{fig:flow} shows the flow diagram of our framework.
The input is represented by RGB and depth frames obtained from a moving RGB-D sensor, which are processed individually.

Our method has four components: SLAM framework, 2D semantic segmentation with a specifically designed CNN, incrementally building a geometric 3D map, and updating class probabilities assigned to each segment of the geometric 3D map.
Firstly, SLAM and semantic segmentation with the CNN are performed simultaneously.
In the segmentation stage, the geometric edge map is generated from the current depth frame and improved with edges extracted from the semantic segmentation result toward the semantic-aware representation.
The geometric 3D map is updated through the edge map, and rendered to the current image plane.
Finally, class probabilities assigned to each segmented region are updated with the rendered segmentation map.
The following section describes these components in more detail.

\subsection{SLAM}
To carry out SLAM in terms of camera pose estimation and fusion we employ the dense approach of InfiniTAM v3 \cite{InfiniTAM_V3_Report_2017}, relying on the efficient and scalable data representation proposed by Keller et al. \cite{keller2013real}, which uses a set \emph{surfels} $s_k$ to build the 3D map.
As per this method, at the $t$-th incoming RGB-D frames, the current camera pose $\bm{T}_t \in \mathbb{SE} (3)$ is estimated through Iterative Closest Point \cite{low2004linear} and RGB alignment.
The new surfels generated from the current depth map are fused into the 3D map by means of the estimated camera pose, and are used to refine the 3D coordinates and normal associated to the existing surfels.

\subsection{CNN architecture}
The details of the CNN architecture proposed in our framework, Low-Res Net, are shown in Fig. \ref{fig:flow} (g).
The architecture combines concepts from state-of-the-art CNN models, i.e. Deep Residual Networks (ResNet) \cite{he2016deep} and FCN  \cite{long2015fully}.
Specifically, the original FCN architecture \cite{long2015fully} utilizes the VGG model \cite{simonyan2014very} to extract features and outputs a semantic segmentation result at the same resolution of the input image.
On the other hand, Low-Res Net employs the ResNet architecture \cite{he2016deep}, which achieved higher accuracy than the VGG model \cite{simonyan2014very} in ImageNet \cite{krizhevsky2012imagenet}, and employs skip connections as done by FCN \cite{long2015fully}.

Towards the goal of achieving a fast forward pass, we do not incorporate multi-layered upsampling and design it only with two deconvolution layers with two strides.
Therefore, given the input image $\mathcal{I}_t(\bm{u}), \bm{u}=(x, y) \subset \mathbb{Z}^2, 0 \leq x < W, 0 \leq y < H$, Low-Res Net outputs a semantic segmentation map in Fig. \ref{fig:flow} (h) as a set of semantic class probabilities, i.e.
\begin{equation} 
\tilde{\mathcal{S}}(\bm{v})=\mathcal{P}(c|\mathcal{I}_t)
\end{equation}
where $\bm{v}=(s, t) \subset \mathbb{Z}^2, 0 \leq s < W/8, 0 \leq t < H/8$. Here, $\mathcal{P}(c)$ denotes a class probability, where $\mathcal{P}(c) \subset \mathbb{R}, 0 \leq \mathcal{P}(c) \leq 1, c \subset \mathbb{Z}, 0 \leq c < N$ with $N$ being the number of categories.
The symbol $\tilde{\ }$ denotes instead hereinafter a map of size $H / 8 \times W / 8$.
In our implementation, $H = 240$, $W = 320$, and the number of channels of the input image $\mathcal{I}$ is 3 as in ResNet \cite{he2016deep}.

\subsection{Segmentation}
Our geometrical segmentation scheme is based on the method proposed by Tateno et al. \cite{tateno2015real}.
The method incrementally builds up a geometric 3D map, where a segmentation label $l_i$ is associated with each surfel $s_k$, by properly propagating and merging segments extracted from the current depth map.

As a result, we obtain a binary geometric edge map $\mathcal{B}^g$ in Fig. \ref{fig:flow} (c) from the input depth frame by comparing neighboring normal angles and vertex distances and by relying on a vertex and normal map as proposed in \cite{tateno2015real}.
Here, $\mathcal{B}^g(\bm{u})$ takes $1$ if $\bm{u}$ is on an edge and $0$ for otherwise.
It is important to point out that, while $ \mathcal{B}^g $ is stable since those edges are extracted geometrically, edges between objects that do not present notable geometric characteristics (e.g., such as with two nearby flat objects) can not be extracted.

\begin{figure*}[t]
\begin{center}
    \includegraphics[width=1.0\hsize]{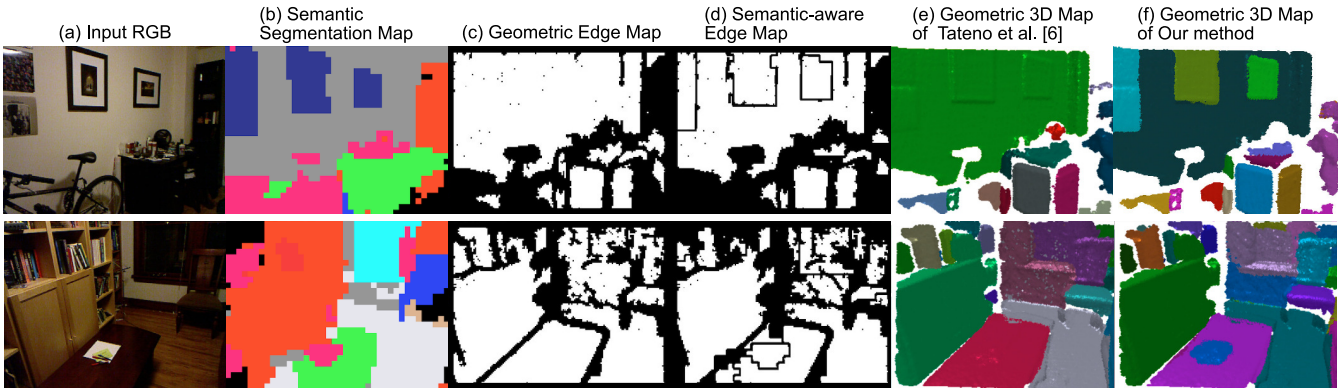}
    \caption{Example results of our segmentation improvement scheme. \label{fig:vsinseg}}
\end{center}
\end{figure*}

Differently from the geometric segmentation from \cite{tateno2015real}, we introduce semantic information into the segments.
First, we generate a class map $\tilde{\mathcal{S}}^c$, where each component $\tilde{\mathcal{S}}^c(\bm{v})$ has a class category $c$, with 
\begin{equation}
\tilde{\mathcal{S}}^c(\bm{v}) = \argmax_{c} \tilde{\mathcal{S}}(\bm{v})=\mathcal{P}(c|\mathcal{I}_t).
\end{equation}

After applying a median filter to $\tilde{\mathcal{S}}^c$ to remove isolated points, we resize $\tilde{\mathcal{S}}^c$ to $\mathcal{S}^c$ with a nearest neighbor interpolation. We would like to point out that the choice of such an efficient interpolation approach over a higher quality resizing such as bilinear interpolation is motivated by the fact that contours of a CNN-based semantic segmentation map are often imprecise, hence a better interpolation method would not yield benefits in terms of accuracy. At the same time, noise in the segment contours is eventually averaged out by the employed confidence-based label fusion approach. Then, we generate a binary semantic edge map $\mathcal{B}^s$ with the following scheme:
\begin{equation}
\label{eq:semedge}
\mathcal{B}^s (\bm{u} = (x, y)) = 
    \left\{%
        \begin{array}{ll}
1 & \mbox{if}\ \ \text{\parbox{0.4\textwidth}{%
$\mathcal{S}^c (x, y) \neq \mathcal{S}^c (x + 1, y) \lor \\ \mathcal{S}^c (x, y) \neq \mathcal{S}^c (x, y + 1) \lor \\ \mathcal{S}^c (x, y) \neq \mathcal{S}^c (x + 1, y + 1)$}} \\
0 & \mbox{otherwise}
        \end{array}%
    \right.
\end{equation}
The final binary semantic-aware edge map $\mathcal{B}$, (d) in Fig. \ref{fig:flow}, is obtained by applying a binary $OR$ operator between $\mathcal{B}^g$ and $\mathcal{B}^s$. 

In Fig. \ref{fig:vsinseg}, the geometric edge map in (c) and the semantic-aware edge map in (d) show the benefit of our segmentation improvement scheme.
Edges between objects which have poor geometric characteristics (i.e., wall and picture in the upper row and desk and paper in the bottom row) are successfully merged to the edge map.

Similar to \cite{tateno2015real}, segments of the semantic-aware edge map $\mathcal{B}$ are properly extracted by means of a connected component algorithm and utilized for incrementally propagating and merging into the geometric 3D map with the estimated camera pose $\bm{T}_t$.

\subsection{Probability fusion}
\label{sec:pf}

Conventional methods assign class probabilities to each element that composes the 3D map \cite{hermans2014dense,mccormac2017semanticfusion,li2016semi,yang2017semantic,vineet2015incremental,kundu2014joint}.
Conversely, we propose to assign class probabilities to each segmentation label $l_i$ associated to each region constituting the geometric 3D map.
With our approach, each label $l_i$ is assigned to a discrete probability distribution $\mathcal{P}(c|\mathcal{I}_{1...t})$ and to a probability confidence $\Gamma$.
$\mathcal{P}(c|\mathcal{I}_{1...t})$ is initialized to $0$ over all class probabilities and $\Gamma$ is also initialized to $0$.
Therefore, the space complexity for storing class probabilities is $O(N \cdot N_l)$, where $N_l$ denotes the number of segmentation labels, in contrast to conventional methods \cite{hermans2014dense,mccormac2017semanticfusion} which require $O(N \cdot N_s)$, where $N_s$ is the number of elements of the 3D map (e.g., the number of surfels). This is an important difference in terms of scalability since typically $N_s \gg N_l$. 
This also appears as a more natural approach, since it could be argued that humans recognize objects by assigning semantic labels in a region-wise manner rather than element-wise.

In order to fuse the output of the Low-Res CNN properly with the 3D map, we update class probabilities assigned to each segmentation label $l_i$ using a confidence-based approach.
Firstly, we render the updated geometric 3D map onto the current image plane using the estimated camera pose $\bm{T}_t$ and the 3D position $\bm{x}(k)$ associated with each surfel $s_k$.
The rendered segmentation map $\mathcal{L}(\bm{u})$, where each component is associated to a segmentation label $l_i$, is generated with $\mathcal{L}(\pi(\bm{T}^{-1}_t \bm{x}(k))) = l_i(k)$ by denoting the segmentation label $l_i$ of a surfel $s_k$ with $l_i (k)$.
Here, $\mathcal{L}(\bm{u})$ takes $\phi$ on the pixel $\bm{u}$ which is not filled with a label $l_i$.

Although the CNN-based semantic segmentation used in our framework is fast, its output $ \tilde{\mathcal{S}} $ has a low resolution.
Using the rendered segmentation map $\mathcal{L}$ whose size is $H \times W$ (i.e. the size of input image), detailed information is introduced to $ \tilde{\mathcal{S}} $ to update the class probabilities of each label $l_i$ with the following update scheme. 

\begin{figure}[t]
\begin{center}
    \includegraphics[width=1.0\hsize]{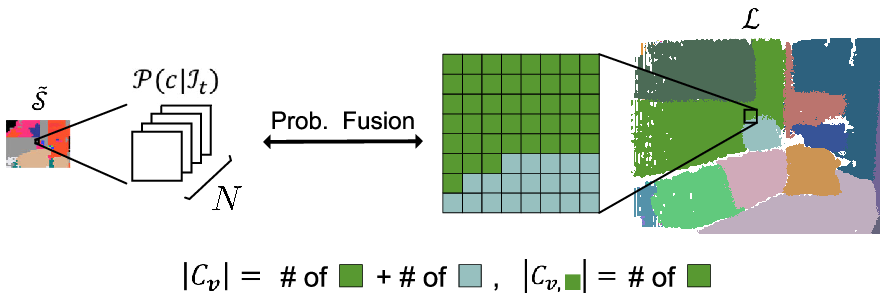}
    \caption{Example definition of set $\mathcal{C}_{\bm{v}}$ and $\mathcal{C}_{\bm{v}, l_i}$. \label{fig:expl}}
\end{center}
\end{figure}

\begin{table*}[t]
   \begin{center}
  \caption{Quantitative results for the NYUv2 dataset \cite{silberman2012indoor}. These results were captured immediately after processing the frame. All accuracy evaluations were performed at $320 \times 240$ resolution. We calculated these accuracies with the same strategies as \cite{mccormac2017semanticfusion}. Ours-Geometric-Only denotes the method of building the geometric 3D map without our segmentation improvement scheme. \label{tab:pre}}

     \scalebox{1.05}{
      \begin{tabular}{l|rrrrrrrrrrrrr|r|r} \bhline{1pt}
         Method & 
         \multicolumn{1}{|c|}{ \cellcolor{bed}\color{white} \rotatebox{90}{bed}} & 
         \multicolumn{1}{c|}{ \cellcolor{books} \rotatebox{90}{books}} & 
         \multicolumn{1}{c|}{ \cellcolor{ceiling}\color{white} \rotatebox{90}{ceiling}} & 
         \multicolumn{1}{c|}{ \cellcolor{chair}\color{white} \rotatebox{90}{chair}} & 
         \multicolumn{1}{c|}{ \cellcolor{floor} \rotatebox{90}{floor}} & 
         \multicolumn{1}{c|}{ \cellcolor{furniture} \rotatebox{90}{furniture}} & 
         \multicolumn{1}{c|}{ \cellcolor{objects} \rotatebox{90}{objects}} & 
         \multicolumn{1}{c|}{ \cellcolor{painting}\color{white} \rotatebox{90}{painting}} & 
         \multicolumn{1}{c|}{ \cellcolor{sofa} \rotatebox{90}{sofa}} & 
         \multicolumn{1}{c|}{ \cellcolor{table} \rotatebox{90}{table}} & 
         \multicolumn{1}{c|}{ \cellcolor{tv} \rotatebox{90}{tv}} & 
         \multicolumn{1}{c|}{ \cellcolor{wall} \rotatebox{90}{wall}} & 
         \multicolumn{1}{c|}{ \cellcolor{window} \rotatebox{90}{window}} & 
         \multicolumn{1}{c|}{ {\bf \rotatebox{90}{class avg.\ }}} & 
         {\bf \rotatebox{90}{pixel avg.\ }} \\ \hline
         Hermans et al. \cite{hermans2014dense} & 68.4 & 45.4 & {\bf 83.4} & 41.9 & 91.5 & 37.1 & 8.6 & 35.8 & 28.5 & 27.7 & 38.4 & 71.8 & 46.1 & 48.0 &54.3 \\ 
         RGBD-SF \cite{mccormac2017semanticfusion} & 61.7 & {\bf 58.5} & 43.4 & 58.4 & 92.6 & 63.7 & {\bf 59.1} & 66.4 & 47.3 & 34.0 & 33.9 & 86.0 & 60.5 & 58.9 &67.5 \\
         RGBD-SF-CRF \cite{mccormac2017semanticfusion} & 62.0 & 58.4 & 43.3 & 59.5 & {\bf 92.7} & 64.4 & 58.3 & 65.8 & 48.7 & 34.3 & 34.3 & 86.3 & 62.3 & 59.2 & 67.9 \\ 
         Eigen-SF \cite{mccormac2017semanticfusion} & 47.8 & 50.8 & 79.0 & 73.3 & 90.5 & 62.8 & 46.7 & 64.5 & 45.8 & 46.0 & 70.7 & 88.5 & 55.2 & 63.2 & 69.3 \\ 
         Eigen-SF-CRF \cite{mccormac2017semanticfusion} & 48.3 & 51.5 & 79.0 & {\bf 74.7} & 90.8 & 63.5 & 46.9 & 63.6 & 46.5 & 45.9 & {\bf 71.5} & {\bf 89.4} & 55.6 & {\bf 63.6} & 69.9 \\ 
         Li et al. \cite{li2016semi} & 64.9 & 34.6 & 72.0 & 67.5 & 90.5 & 65.0 & 17.2 & {\bf 67.3} & 59.3 & 41.3 & 60.0 & 85.1 & 57.0 & 60.3 & 70.3 \\ \hline
         Ours-Geometric-Only & {\bf 83.7} & 6.4 & 32.0 & 52.8 & 83.1 & 73.5 & 40.0 & 4.3 & 75.3 & {\bf 56.6} & 53.1 & 75.0 & 50.2 & 52.8 & 66.9 \\
         Ours & {\bf 83.7} & 15.6 & 24.4 & 56.7 & 83.3 & {\bf 76.1} & 52.5 & 40.8 & {\bf 77.7} & 53.0 & 57.3 & 75.3 & {\bf 64.4} & 58.5 & {\bf 70.7} \\ \bhline{1pt}
      \end{tabular}
      }
   \end{center}
\end{table*}

First, a set $\mathcal{C}_{\bm{v}}$ and a set $\mathcal{C}_{\bm{v}, l_i}$ are defined as
\begin{equation}
\label{eq:set_cv}
\begin{split}
\mathcal{C}_{\bm{v}=(s, t)} = \bigl\{ \bm{u} = (x, y) \subset \mathbb{Z}^2 | \mathcal{L}(\bm{u}) \neq \phi \land \\ 8s \leq x < 8(s + 1) \land 8t \leq y < 8(t + 1) \bigr\}
\end{split}
\end{equation}
and
\begin{equation}
\label{eq:set_cvli}
\begin{split}
\mathcal{C}_{\bm{v}, l_i} = \bigl\{ \bm{u} \subset \mathcal{C}_{\bm{v}} | \mathcal{L}(\bm{u}) = l_i \bigr\} \mbox{.}
\end{split}
\end{equation}
In words: $\mathcal{C}_{\bm{v}}$ is a set of coordinates to which the labels are assigned in the region of $\mathcal{L}(\bm{u})$ corresponding to $\tilde{\mathcal{S}}(\bm{v})$, while $\mathcal{C}_{\bm{v}, l_i}$ is a set of coordinates to which the label $l_i$ is assigned (See Fig. \ref{fig:expl}).

When the set $\mathcal{U}_{\bm{v}}$ of labels $l_i$ which is included in the region of $\mathcal{L}(\bm{u})$ corresponding to $\tilde{\mathcal{S}}(\bm{v})$ is defined as
\begin{equation}
\label{eq:set_label}
\begin{split}
\mathcal{U}_{\bm{v} = (s, t)} = \bigl\{ l_i = \mathcal{L}(x, y) \subset \mathbb{Z} | 8s \leq x < 8(s+1) \land \\ 8t \leq y < 8(t+1) \bigr\} \mbox{,}
\end{split}
\end{equation}
the class probabilities $\mathcal{P}(c|\mathcal{I}_{1...t})$ and the probability confidence $\Gamma$ of each element $l \subset \mathcal{U}_{\bm{v}}$ are updated through
\begin{equation}
\label{eq:update}
\begin{split}
\mathcal{P}(c|\mathcal{I}_{1...t}) \gets \frac{1}{Z} \cdot \frac{\Gamma \mathcal{P}(c|\mathcal{I}_{1...t-1}) + {\gamma} \mathcal{P}(c|\mathcal{I}_{t})}{\Gamma + \gamma} \\
\Gamma \gets \Gamma + \gamma, \  \gamma = \frac{| \mathcal{C}_{\bm{v}, l} |}{| \mathcal{C}_{\bm{v}} |}
\end{split}
\end{equation}
which is applied to all class probabilities.
Here, the constant $Z$ is for normalizing class probabilities to a proper distribution.
With this scheme, the weight of the probability which cross over two or more segment regions (e.g., wall and object in Fig. \ref{fig:expl}) is reduced.
By applying the same strategy to all $ \bm{v} $ constituting $ \tilde{\mathcal{S}}(\bm{v}) $, we update class probabilities of all labels included in the rendered segmentation map $\mathcal{L}(\bm{u})$.

Therefore, letting the size of $ \tilde{\mathcal{S}} (\bm{v}), H/8 \times W/8 $ be $\tilde{H} \times \tilde{W}$, the time complexity for updating class probabilities is $\mathcal{O}( \tilde{H} \tilde{W} (8 \times 8 + | \mathcal{U}_{\bm{v}} | N))$, which means calculating set $\mathcal{C}_{\bm{v}}$, $\mathcal{C}_{\bm{v}, l_i}$, and $\mathcal{U}_{\bm{v}}$ takes $8 \times 8$ and updating all class probabilities $N$ assigned to each label in $\mathcal{U}_{\bm{v}}$ takes $|\mathcal{U}_{\bm{v}}|N$.
Note that conventional methods \cite{hermans2014dense,mccormac2017semanticfusion,yang2017semantic,vineet2015incremental} take $\mathcal{O}(HWN)$ for updating class probabilities of the 3D map with a frame-wise recognition.

\section{EXPERIMENTS}

\subsection{Dataset and implementation}
We evaluate our system on the common NYUv2 dataset \cite{silberman2012indoor}.
The dataset contains 206 test set video sequences, however, for a fair comparison, we picked up 140 test sequences having a frame-rate over 2Hz which is the same as \cite{mccormac2017semanticfusion}.
Since our Low-Res CNN outputs semantic segmentation with the size of $W/8 \times H/8$, we resized the ground truth $\mathcal{S}_{gt}$ to $\tilde{\mathcal{S}}_{gt}$ by filling $\tilde{\mathcal{S}}_{gt} (\bm{v})$ with the label which mostly occupies the area of $\mathcal{S}_{gt} (\bm{u})$ corresponding to $\tilde{\mathcal{S}}_{gt} (\bm{v})$.
After training our Low-Res Net with the MS COCO dataset \cite{lin2014microsoft} for 10 epochs, we fine-tuned the network with the training dataset of the NYUv2 dataset \cite{silberman2012indoor} for 50 epochs.
These evaluations are conducted on an Intel Core i7-5557U 3.1GHz CPU, GeForce GTX 1080 GPU, and 16GB RAM.

\subsection{Accuracy}
\label{sec:res}

\begin{figure*}[t]
\begin{center}
    \includegraphics[width=1.0\hsize]{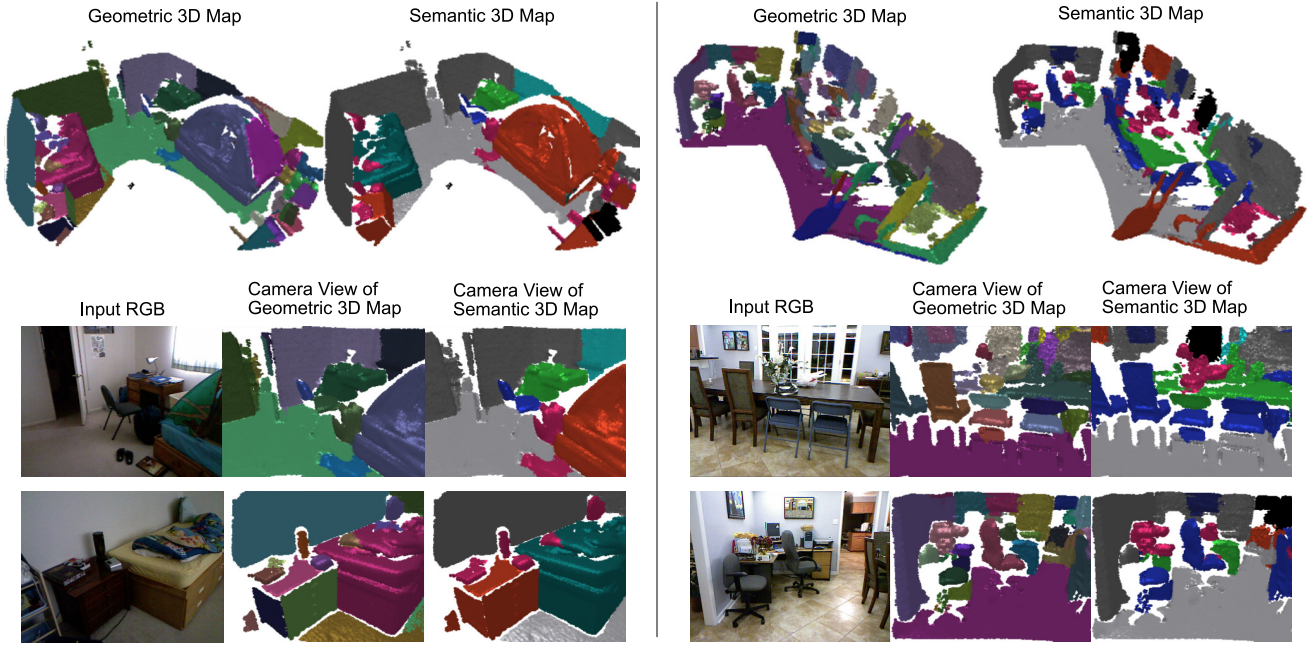}
    \caption{Qualitative results of our dense 3D semantic mapping on two scenes (left: {\it bedroom\_0112}, right: {\it dining\_room\_0017}). See Table \ref{tab:pre} for class colors. \label{fig:map}}
\end{center}
\end{figure*}

\begin{figure*}[t]
\begin{center}
    \includegraphics[width=1.0\hsize]{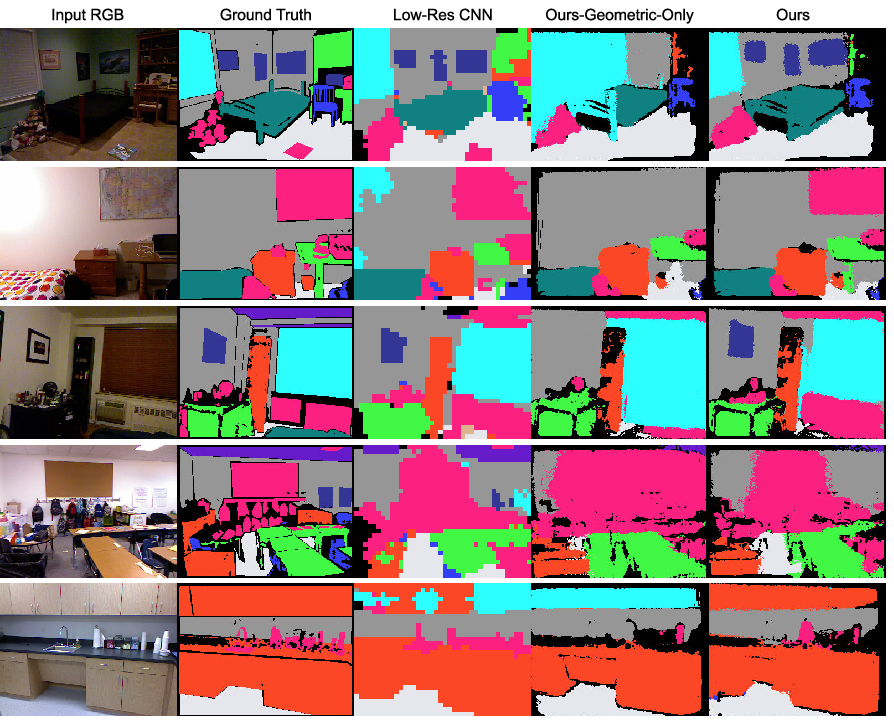}
    \caption{Qualitative results for the NYUv2 dataset \cite{silberman2012indoor}. As with Table \ref{tab:pre}, Ours-Geometric-Only denotes the method of building the geometric 3D map without our segmentation improvement scheme. See Table \ref{tab:pre} for class colors. \label{fig:comp}}
\end{center}
\end{figure*}

In this section, we experimentally demonstrate the accuracy of our method by quantitatively comparing the accuracy with other state-of-the-art methods through Table \ref{tab:pre}. Additionally, Fig. \ref{fig:map} and Fig. \ref{fig:comp} show qualitative results of our dense semantic mapping.

As shown in Table \ref{tab:pre}, our method achieves 0.8\% higher average pixel accuracy compared to SemanticFusion \cite{mccormac2017semanticfusion} and 0.4\% higher average pixel accuracy compared to Li et al. \cite{li2016semi}.
As it can be noted, our method is particularly capable of outperforming other semantic mapping methods for object categories characterized by a big size. 
For the class {\it bed}, there is a significant accuracy increase of 15.3\% over the state of the art; while, for the class {\it furniture} and {\it sofa}, we achieve 11.1\% and 18.4\% improvement, respectively.
The reason why we achieve high accuracy especially on such categories is that our segmentation strongly relies on geometric information, and geometric boundaries associated to these categories (e.g., {\it bed} and {\it wall} and {\it floor} and {\it furniture}) are often quite clear.

Fig. \ref{fig:comp} shows the benefit of the segmentation improvement from the viewpoint of accuracy compared with ``Ours-Geometric-Only'', where we build the geometric 3D map without our segmentation improvement scheme.
Particularly in the upper three rows, the {\it paintings} and the {\it window} on the {\it wall}, which are difficult to distinguish only with the geometric-based segmentation, are also segmented and annotated correctly.
The geometric 3D map in Fig. \ref{fig:map} also shows the validity of the segmentation improvement especially on the above-mentioned regions.
The example results of building a geometric 3D map with/without segmentation improvement are in Fig. \ref{fig:vsinseg} (e) geometric 3D map of Tateno et al. \cite{tateno2015real} and (f) geometric 3D map of our method.
We achieved semantic-aware representation rather than the geometric-only incremental segmentation method \cite{tateno2015real}.
This improved segmentation scheme allows achieving higher accuracy in terms of pixel average than state-of-the-art methods.
As shown in Table \ref{tab:pre}, the accuracies of the class {\it painting} and {\it window} are significantly improved for 36.5\% and 14.2\%, respectively, and 3.8\% for overall categories between ``Ours'' and ``Ours-Geometric-Only''.

The lower two rows of Fig. \ref{fig:comp} show failure cases.
Since our method mainly extracts edges from the vertex and normal map obtained from the incoming depth image, it is difficult to successfully segment distant objects where depth values tend to be unstable (i.e., the third row of Fig. \ref{fig:comp}) and manage scenes where many small objects are lined up where vertices and normals are cluttered (i.e., the fourth row of Fig. \ref{fig:comp}).
In Table \ref{tab:pre}, this is the same reason why the categories of small objects such as {\it book} and {\it objects} score low accuracies.
We leave the exploration of improving these limitation to future work.

\subsection{Computational cost}
\label{sec:exp_cc}

\begin{table}[tb]
   \begin{center}
  \caption{Comparison of run-time performance. FQ denotes the frequency to perform a recognition of the input frame and update class probabilities of the 3D map. \label{tab:runtime}}
       \scalebox{1.0}{
      \begin{tabular}{lrrr} \bhline{1pt}
      Method & 3D map &  FQ & FPS \\ \hline 
      Hermans et al. \cite{hermans2014dense} & Dense & every 6 frames & 3.9 - 4.6 Hz \\ 
      SemanticFusion \cite{mccormac2017semanticfusion} & Dense & every 10 frames & 25.3 Hz \\ 
      Yang et al. \cite{yang2017semantic} & Dense & every frame & 2 Hz \\
      Li et al. \cite{li2016semi} & Semi-Dense & every key-frame & 10 Hz \\ \hline 
      Ours & Dense & every frame & {\bf 30.9 Hz} \\ \bhline{1pt}
      \end{tabular}
      }
    \end{center}
\end{table}

\begin{table}[tb]
   \begin{center}
  \caption{Average time spent on each processing stage. processing for segmentation are in line 2-4 and processing for recognition are in 5-7. Note that the processing with * and the processing with ** can be processed simultaneously. \label{tab:ana}}
     \scalebox{1.0}{
       \begin{tabular}{lr} \bhline{1pt}
 Component & Consumed time \\ \hline
 SLAM * & 8.13 ms \\ \hline
 Generate a binary geometric edge map $\mathcal{B}^g$ * & 1.04 ms\\ 
 Segmentation improvement & 0.39 ms \\ 
 Update the geometric 3D map & 8.74 ms \\ \hline
 Low-Res CNN ** & 19.32 ms \\ 
 Generate a rendered segmentation map $\mathcal{L}$ & 2.52 ms \\ 
 Probability fusion & 1.37 ms \\ \hline 
 Total & 32.34 ms \\ \bhline{1pt}
 \end{tabular}
 }
    \end{center}
\end{table}

\begin{figure}[t]
\begin{center}
    \includegraphics[width=1.0\hsize]{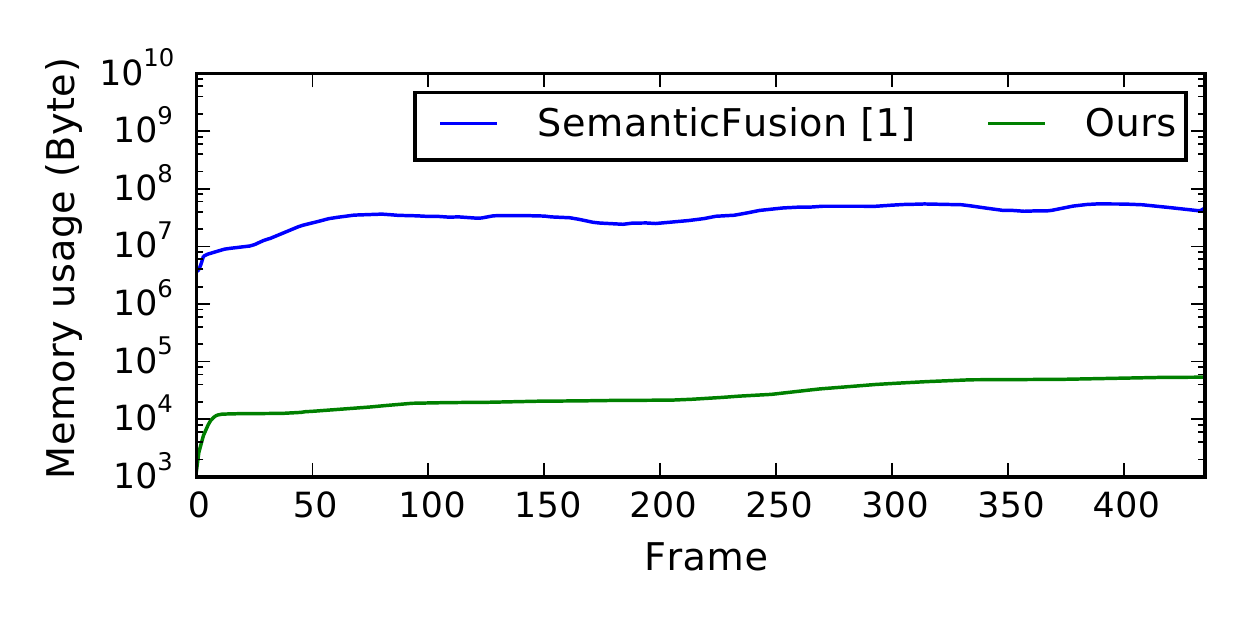}
    \caption{Comparison of memory usage for storing class probabilities with SemanticFusion \cite{mccormac2017semanticfusion} on the scene {\it bedroom\_0112} of the NYUv2 dataset \cite{silberman2012indoor}. \label{fig:space}}
\end{center}
\end{figure}

In this section, we demonstrate the advantage of reducing the computational complexity, i.e. one of the main contributions of this method.
We quantitatively compare the run-time performance with state-of-the-art approaches through Table \ref{tab:runtime}.

As shown in Table \ref{tab:runtime}, we achieved real-time performance (i.e., over 30Hz) while performing all processing components on every input frame.
As analyzed in the last paragraph of Sec. \ref{sec:pf}, the time complexity for updating class probabilities of the 3D map (i.e., Probability fusion) is $\mathcal{O}( \tilde{H} \tilde{W} (8 \times 8 + | \mathcal{U}_{\bm{v}} | N))$.
Considering the average number of $| \mathcal{U}_{\bm{v}} |$ was $1.28$ through the experiments, the average time complexity $\mathcal{O}( \tilde{H} \tilde{W} (8 \times 8 + | \mathcal{U}_{\bm{v}} | N))$ turns into $\mathcal{O}( \tilde{H} \tilde{W} (8 \times 8 + N)) = \mathcal{O}(HW + \tilde{H} \tilde{W} N)$ in contrast to the one of conventional methods $\mathcal{O}(HWN)$ \cite{hermans2014dense,mccormac2017semanticfusion,yang2017semantic,vineet2015incremental}.
Therefore, as shown in Table \ref{tab:ana}, updating class probabilities of the 3D map only took 1.37ms on average, whereas SemanticFusion \cite{mccormac2017semanticfusion} spent 41.1ms for the processing.
Furthermore, the processing for 2D recognition (i.e., Low-Res CNN) only took 19.32ms while maintaining high accuracy in the end, as mentioned in Section. \ref{sec:res}.

Lastly, we discuss about the results of reducing the space complexity through Fig. \ref{fig:space}.
As shown there, the memory usage of our method is significantly reduced compared to the one of SemanticFusion \cite{mccormac2017semanticfusion} over all frames.
The average memory usage of our method is 0.08\% of the one of SemanticFusion \cite{mccormac2017semanticfusion}.
The reason for this significant improvement is that, as mentioned in Sec. \ref{sec:pf}, the space complexity of our method is $O(N \cdot N_l)$ whereas SemanticFusion takes $O(N \cdot N_s)$, where $N_l$ and $N_s$ were $1032$ and $844260$ in the end of the scene respectively.

\section{CONCLUSION}

In this paper, we proposed an efficient semantic mapping approach by assigning class probabilities to each region of the geometric 3D map which is incrementally built up through a robust SLAM framework and a geometric-based incremental segmentation.
Through our experiments, we demonstrated that our approach notably compressed the computational complexity in terms of both of time and space while achieving comparable accuracy against state-of-the-art approaches without any post-processing to the semantic 3D map.
Furthermore, we confirmed that our strategy improved the incremental segmentation framework beyond the geometric only to the semantic-aware representation.

\section*{ACKNOWLEDGMENT}
This research presentation is supported in part by a research assistantship of a Grant-in-Aid to the Program for Leading Graduate School for ``Science for Development of Super Mature Society'' from the Ministry of Education, Culture, Sport, Science, and Technology in Japan.

\bibliographystyle{ieeetr}
\bibliography{template}

\begin{thebibliography}{10}

\bibitem{mccormac2017semanticfusion}
J.~McCormac, A.~Handa, A.~Davison, and S.~Leutenegger, ``Semanticfusion: Dense
  3d semantic mapping with convolutional neural networks,'' in {\em IEEE
  International Conference on Robotics and Automation (ICRA)}, pp.~4628--4635,
  IEEE, 2017.

\bibitem{tateno2015real}
K.~Tateno, F.~Tombari, and N.~Navab, ``Real-time and scalable incremental
  segmentation on dense slam,'' in {\em IEEE/RSJ International Conference on
  Intelligent Robots and Systems (IROS)}, pp.~4465--4472, IEEE, 2015.

\bibitem{li2016semi}
X.~Li and R.~Belaroussi, ``Semi-dense 3d semantic mapping from monocular
  slam,'' {\em arXiv preprint arXiv:1611.04144}, 2016.

\bibitem{yang2017semantic}
S.~Yang, Y.~Huang, and S.~Scherer, ``Semantic 3d occupancy mapping through
  efficient high order crfs,'' {\em IEEE/RSJ International Conference on
  Intelligent Robots and Systems (IROS)}, 2017.

\bibitem{hermans2014dense}
A.~Hermans, G.~Floros, and B.~Leibe, ``Dense 3d semantic mapping of indoor
  scenes from rgb-d images,'' in {\em IEEE International Conference on Robotics
  and Automation (ICRA)}, pp.~2631--2638, IEEE, 2014.

\bibitem{InfiniTAM_V3_Report_2017}
V.~A. Prisacariu, O.~K{\"a}hler, S.~Golodetz, M.~Sapienza, T.~Cavallari, P.~H.
  Torr, and D.~W. Murray, ``{InfiniTAM v3: A Framework for Large-Scale 3D
  Reconstruction with Loop Closure},'' {\em ArXiv e-prints}, 2017.

\bibitem{keller2013real}
M.~Keller, D.~Lefloch, M.~Lambers, S.~Izadi, T.~Weyrich, and A.~Kolb,
  ``Real-time 3d reconstruction in dynamic scenes using point-based fusion,''
  in {\em International Conference on 3DTV-Conference}, pp.~1--8, IEEE, 2013.

\bibitem{kundu2014joint}
A.~Kundu, Y.~Li, F.~Dellaert, F.~Li, and J.~M. Rehg, ``Joint semantic
  segmentation and 3d reconstruction from monocular video,'' in {\em European
  Conference on Computer Vision (ECCV)}, pp.~703--718, Springer, 2014.

\bibitem{silberman2012indoor}
N.~Silberman, D.~Hoiem, P.~Kohli, and R.~Fergus, ``Indoor segmentation and
  support inference from rgbd images,'' in {\em European Conference on Computer
  Vision (ECCV)}, pp.~746--760, Springer, 2012.

\bibitem{vineet2015incremental}
V.~Vineet, O.~Miksik, M.~Lidegaard, M.~Nie{\ss}ner, S.~Golodetz, V.~A.
  Prisacariu, O.~K{\"a}hler, D.~W. Murray, S.~Izadi, P.~P{\'e}rez, {\em
  et~al.}, ``Incremental dense semantic stereo fusion for large-scale semantic
  scene reconstruction,'' in {\em IEEE International Conference on Robotics and
  Automation (ICRA)}, pp.~75--82, IEEE, 2015.

\bibitem{noh2015learning}
H.~Noh, S.~Hong, and B.~Han, ``Learning deconvolution network for semantic
  segmentation,'' in {\em IEEE Conference on Computer Vision and Pattern
  Recognition (CVPR)}, pp.~1520--1528, 2015.

\bibitem{sengupta2013urban}
S.~Sengupta, E.~Greveson, A.~Shahrokni, and P.~H. Torr, ``Urban 3d semantic
  modelling using stereo vision,'' in {\em IEEE International Conference on
  Robotics and Automation (ICRA)}, pp.~580--585, IEEE, 2013.

\bibitem{koppula2011semantic}
H.~S. Koppula, A.~Anand, T.~Joachims, and A.~Saxena, ``Semantic labeling of 3d
  point clouds for indoor scenes,'' in {\em Advances in neural information
  processing systems}, pp.~244--252, 2011.

\bibitem{zhao2016building}
Z.~Zhao and X.~Chen, ``Building 3d semantic maps for mobile robots using rgb-d
  camera,'' {\em Intelligent Service Robotics}, vol.~9, no.~4, pp.~297--309,
  2016.

\bibitem{salas2013slam++}
R.~F. Salas-Moreno, R.~A. Newcombe, H.~Strasdat, P.~H. Kelly, and A.~J.
  Davison, ``Slam++: Simultaneous localisation and mapping at the level of
  objects,'' in {\em IEEE Conference on Computer Vision and Pattern Recognition
  (CVPR)}, pp.~1352--1359, IEEE, 2013.

\bibitem{bowman2017probabilistic}
S.~L. Bowman, N.~Atanasov, K.~Daniilidis, and G.~J. Pappas, ``Probabilistic
  data association for semantic slam,'' in {\em IEEE International Conference
  on Robotics and Automation (ICRA)}, pp.~1722--1729, IEEE, 2017.

\bibitem{galvez2016real}
D.~G{\'a}lvez-L{\'o}pez, M.~Salas, J.~D. Tard{\'o}s, and J.~Montiel,
  ``Real-time monocular object slam,'' {\em Robotics and Autonomous Systems},
  vol.~75, pp.~435--449, 2016.

\bibitem{fioraio2013joint}
N.~Fioraio and L.~Di~Stefano, ``Joint detection, tracking and mapping by
  semantic bundle adjustment,'' in {\em IEEE Conference on Computer Vision and
  Pattern Recognition (CVPR)}, pp.~1538--1545, IEEE, 2013.

\bibitem{long2015fully}
J.~Long, E.~Shelhamer, and T.~Darrell, ``Fully convolutional networks for
  semantic segmentation,'' in {\em IEEE Conference on Computer Vision and
  Pattern Recognition (CVPR)}, pp.~3431--3440, 2015.

\bibitem{badrinarayanan2017segnet}
V.~Badrinarayanan, A.~Kendall, and R.~Cipolla, ``Segnet: A deep convolutional
  encoder-decoder architecture for image segmentation,'' {\em IEEE transactions
  on pattern analysis and machine intelligence (TPAMI)}, vol.~39, no.~12,
  pp.~2481--2495, 2017.

\bibitem{chen2017rethinking}
L.-C. Chen, G.~Papandreou, F.~Schroff, and H.~Adam, ``Rethinking atrous
  convolution for semantic image segmentation,'' {\em arXiv preprint
  arXiv:1706.05587}, 2017.

\bibitem{Uckermann2013}
A.~Uckermann, R.~Haschke, and H.~Ritter, ``{Realtime 3D segmentation for
  human-robot interaction},'' in {\em 2013 IEEE/RSJ International Conference on
  Intelligent Robots and Systems (IROS)}, 2013.

\bibitem{Uckermann2012}
A.~Uckermann, C.~Elbrechter, R.~Haschke, and H.~Ritter, ``{3D scene
  segmentation for autonomous robot grasping},'' in {\em 2012 IEEE/RSJ
  International Conference on Intelligent Robots and Systems (IROS)}, oct 2012.

\bibitem{Pieropan2014}
A.~Pieropan and H.~Kjellstrom, ``{Unsupervised object exploration using
  context},,'' in {\em The 23rd IEEE International Symposium on Robot and Human
  Interactive Communication (RO-MAN)}, 2014.

\bibitem{Abramov2012}
A.~Abramov, K.~Pauwels, J.~Papon, F.~Worgotter, and B.~Dellen,
  ``Depth-supported real-time video segmentation with the kinect,'' in {\em
  IEEE Workshop on Applications of Computer Vision (WACV)}, 2012.

\bibitem{Finman2013}
R.~Finman, T.~Whelan, M.~Kaess, and J.~J. Leonard, ``{Toward lifelong object
  segmentation from change detection in dense RGB-D maps},'' in {\em 2013
  European Conference on Mobile Robots, ECMR 2013 - Conference Proceedings},
  pp.~178--185, 2013.

\bibitem{low2004linear}
K.-L. Low, ``Linear least-squares optimization for point-to-plane icp surface
  registration,'' {\em Chapel Hill, University of North Carolina}, vol.~4,
  2004.

\bibitem{he2016deep}
K.~He, X.~Zhang, S.~Ren, and J.~Sun, ``Deep residual learning for image
  recognition,'' in {\em Proceedings of the IEEE conference on computer vision
  and pattern recognition (CVPR)}, pp.~770--778, 2016.

\bibitem{simonyan2014very}
K.~Simonyan and A.~Zisserman, ``Very deep convolutional networks for
  large-scale image recognition,'' {\em arXiv preprint arXiv:1409.1556}, 2014.

\bibitem{krizhevsky2012imagenet}
A.~Krizhevsky, I.~Sutskever, and G.~E. Hinton, ``Imagenet classification with
  deep convolutional neural networks,'' in {\em Advances in neural information
  processing systems (NIPS)}, pp.~1097--1105, 2012.

\bibitem{lin2014microsoft}
T.-Y. Lin, M.~Maire, S.~Belongie, L.~Bourdev, R.~Girshick, J.~Hays, P.~Perona,
  D.~Ramanan, C.~L. Zitnick, and P.~Dollar, ``Microsoft coco: Common objects in
  context,'' {\em arXiv preprint arXiv:1405.0312}, 2014.

\end{thebibliography}
\end{document}